\documentclass[conference]{IEEEtran}
\IEEEoverridecommandlockouts
\usepackage{cite}
\usepackage{amsmath,amssymb,amsfonts}
\usepackage{algorithmic}
\usepackage{graphicx}
\usepackage{textcomp}
\usepackage{xcolor}
\usepackage{authblk}
\usepackage{multirow}
\usepackage{makecell}
\usepackage{bbm}
\usepackage{dsfont}
\def\BibTeX{{\rm B\kern-.05em{\sc i\kern-.025em b}\kern-.08em
    T\kern-.1667em\lower.7ex\hbox{E}\kern-.125emX}}
\begin{document}{


\title{EasyVis2: A Real-Time Multi-view 3D Visualization System for Laparoscopic Surgery Training Enhanced by a Deep Neural Network YOLOv8-Pose
\thanks{This work was supported by the National Institute of Biomedical Imaging and Bioengineering (NIBIB) of the U.S. National Institutes of Health (NIH) under award number R01EB019460.}

}
\author{%
Yung-Hong Sun$^{1*}$ \quad Gefei Shen$^1$ \quad Jiangang Chen$^1$ \quad Jayer Fernandes$^1$\\
Amber L. Shada$^2$ \quad Charles P. Heise$^2$ \quad Hongrui Jiang$^{1}$ \quad Yu Hen Hu$^{1}$\\
$^1$Department of Electrical and Computer Engineering, University of Wisconsin - Madison, WI 53705, USA \\
$^2$Department of Surgery, University of Wisconsin - Madison, WI 53705, USA \\
$^*$ysun376@wisc.edu
}

}

\maketitle

\begin{abstract}
EasyVis2 is a system designed to provide hands-free, real-time 3D visualization for laparoscopic surgery. It incorporates a surgical trocar equipped with an array of micro-cameras, which can be inserted into the body cavity to offer an enhanced field of view and a 3D perspective of the surgical procedure. A specialized deep neural network algorithm, YOLOv8-Pose, is utilized to estimate the position and orientation of surgical instruments in each individual camera view. These multi-view estimates enable the calculation of 3D poses of surgical tools, facilitating the rendering of a 3D surface model of the instruments, overlaid on the background scene, for real-time visualization. This study presents methods for adapting YOLOv8-Pose to the EasyVis2 system, including the development of a tailored training dataset. Experimental results demonstrate that, with an identical number of cameras, the new system improves 3D reconstruction accuracy and reduces computation time. Additionally, the adapted YOLOv8-Pose system shows high accuracy in 2D pose estimation.

\end{abstract}


\section{Introduction}
Laparoscopic Surgery (LS) is performed through small incisions in the body cavity using specialized surgical tools and a laparoscope\cite{article,inproceedings,10.1007/978-3-540-75757-3_9,10.1007/978-3-031-16449-1_41}. One significant challenge encountered is the inherent difficulty in perceiving three-dimensional (3D) depth when viewing the abdominal cavity on a two-dimensional (2D) monitor\cite{son2023advancements}. This limitation can potentially impact the precision and safety of the procedure. A common method to perceive 3D depth from a 2D view is to shift perspective or change the viewing angle\cite{grossberg2021conscious}. Current methods for sensing 3D involve a human assistant manually operating the camera during surgery\cite{katz2022dual}.

Maneuver-free 3D visualization algorithms have been developed to address this challenge, enabling the viewing of the body cavity from novel angles without manually maneuvering the endoscope. Early works\cite{10.1007/978-3-540-75757-3_9,ke2020towards,article,inproceedings,ackerman2002surface,maurice2012structured,clancy2015dual,reiter2014surgical} rely on time-consuming algorithms based on classical feature points\cite{ghahremani2020ffd,bouguet1995proceedings,rosten2005fusing} that require distinguishable textures and cannot work on non-textured surgical tools. A Neural Radiance Field (NeRF)\cite{mildenhall2021nerf} based method \cite{10.1007/978-3-031-16449-1_41} was proposed to solve the laparoscopic surgery (LS) 3D rendering challenge. However, it ignored the surgical tools, and the slow frame-wise learning speed and the requirement for dense views in NeRF posed significant challenges to its application in real surgery. 

EasyVis \cite{sun2025easyvis} was developed to solve the real-time 3D rendering challenge for laparoscopic surgery under the LS box trainer bean drop task constraint. It processes moving and static objects separately and utilizes Augmented Reality (AR) techniques to achieve real-time rendering for moving rigid surgical tools by estimating the surgical tool 3D pose and then completing the virtual surgical tool surface model according to the pose. The object pose describes its state in the 2D or 3D space, such as its position and direction. However, it relies on color markers to estimate the 3D object pose. The color detection in this framework is sensitive to environmental light sources and can be easily influenced by other colors in the system, limiting its use to laparoscopic surgery box trainers with a pre-determined background.

We proposed this work to address these limitations. We used YOLOv8-Pose \cite{yolov8} as the 2D object pose estimator to improve the original EasyVis. The YOLO series was used in a variety of fields \cite{algiriyage2021towards,qureshi2023comprehensive,xu2024surgical,almufareh2024automated,maji2022yolo,rahati2022sports}, inspired by these applications and due to its efficiency, we chose YOLOv8-Pose, the latest model at the time we started this work, to estimate the surgical tool skeleton and integrate it into our EasyVis pipeline. 


\begin{figure*}[!t]
    \centering
    \includegraphics[width=\linewidth]{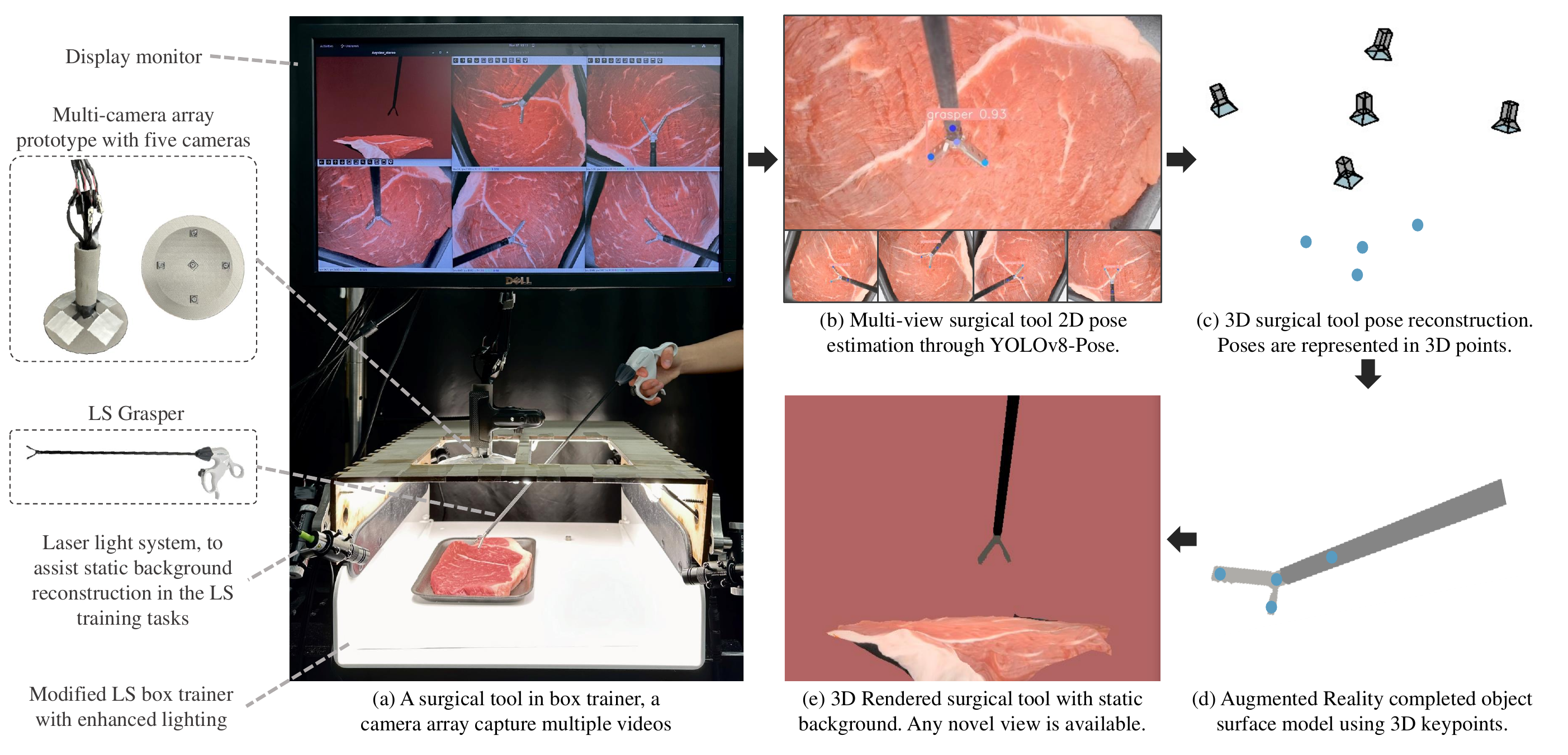}
    \caption{Real-time 3D Rendering framework for surgical tool. a) Laparoscopic surgery tool in the laparoscopic surgery box trainer, a camera array with five cameras captures the video stream. b) Implement surgical tool 2D skeleton estimation for each camera view and each video frame through YOLOv8-Pose. c) Estimate surgical tool 3D skeleton through 3D reconstruction. d) Use Augmented Reality to complete the object surface model. e) Render the reconstructed 3D skeleton with a virtual surgical tool model to any view.}
    \label{workflow}
\end{figure*}

Surgical tools such as graspers, when viewed under a laparoscope, can have their pose simplified into four points to describe their states in a 2D image or 3D space: two points for the tip ends, one for the wrist, and one point to assist in determining the grasper’s direction. The first three points are well-founded and can be easily defined. However, defining the fourth point presents a challenge due to the lack of a clear basis. To address this issue, we proposed a novel training technique. We introduced a marker to help locate this point during data labeling and then use a data augmentation approach to remove the model’s dependence on the marker, achieving marker-free consistent pose estimation.

Existing surgical tool datasets primarily focus on segmentation or classification tasks \cite{sznitman2012data, maier2014can}, but none describe the surgical tool skeleton, particularly for laparoscopic surgical tools that are only partially visible in the view. To address this gap, we propose a surgical tool pose dataset (ST-Pose). This dataset enables the development of a feasible surgical tool pose estimation model for 3D reconstruction. A deep neural network (DNN) model typically requires a substantial amount of data to achieve high accuracy, but the exact quantity needed remains unknown due to the black-box nature of neural networks. Sampling data under the guidance of a model trained with an existing dataset can efficiently maximize data quality. By sampling data points with the highest error rates, we can compensate for deficiencies in the dataset. This approach allows us to estimate the amount of data required to train a high-quality model and make the dataset representative enough. Additionally, generating a dataset is labor-intensive and requires significant human resources to cover all potential conditions. To maximize efficiency with our limited labor force, we sample our data in a controlled environment and then use data synthesis techniques to enable the model to function in various environments.

This work extends the capabilities of EasyVis, allowing it to work on marker-free rigid surgical tools in the LS box trainer environment, with the potential to be applied in real surgery. In this paper, we introduce a Surgical Tool Pose dataset (ST-Pose) to train a 2D object skeleton estimation model YOLOv8-Pose. A DNN training strategy is introduced to remove the dependence on markers, achieving marker-free surgical tool pose estimation with good robustness. The dataset includes captured surgical tool images under LS view with labeled poses, object masks, and marker masks. It specifically covers a type of LS grasper and the beans used in the LS bean drop task, aiming to achieve robust prediction in the bean drop task of LS training. Additionally, an LS scissor is included in the dataset as an extension. We employed semi-automatic methods to generate the dataset efficiently. A trained YOLOv8-Pose model is deployed in the EasyVis framework with TensorRT GPU optimization to improve efficiency. Experiments were performed to demonstrate the precision of our approach and its potential for real surgery. The improved EasyVis framework is shown in Fig. \ref{workflow}. This work demonstrates a novel idea that incorporates 2D LS tool pose estimation with 3D reconstruction and rendering in LS training, paving the way for applications in 3D visualization during actual surgery.

\section{Related Work}
\subsection{3D Rigid Object Pose Estimation}

3D object pose estimation is a crucial intermediate result in our system pipeline. Existing algorithms are divided into end-to-end methods and multi-stage methods. End-to-end methods directly generate 3D poses as output. Some of these methods only require RGB images as input \cite{zhao20206d}, \cite{zhu2021aspp}, \cite{wang2021gdr}, while others use RGB-D, requiring additional depth maps generated from either depth cameras or lidar \cite{wen2024foundationpose}, \cite{9197555}, \cite{song2024rgb}. These methods typically require a significant amount of training data. Our system faces challenges in aligning camera coordinates under a multi-camera setup, and generating an RGB-D dataset with labeled ground truth using existing depth cameras may be difficult due to the short range of our setup. Moreover, these methods only estimate 3D poses in 6 degrees of freedom (3D position and rotation), whereas the surgical tool states, such as the tip's open angle, are additionally required. Therefore, we decided to retain the EasyVis framework, which uses a multi-stage pipeline to estimate the object’s 3D pose. This framework first estimates the 2D pose from each image and then estimates the 3D pose using 2D poses from multiple views through 3D triangulation \cite{hartley2003multiple}.

Existing 2D pose estimation methods typically use a set of keypoints to define the object’s structure. \textit{DeepPose} \cite{toshev2014deeppose} is one of the pioneering methods that introduced deep neural networks for pose estimation, using a two-stage approach to estimate poses from coarse to fine. \textit{OpenPose} \cite{openpose} and \textit{DeepCut} \cite{pishchulin2016deepcut} employ bottom-up methods to achieve fast and highly accurate pose detection. \textit{DeepCut} detects a set of keypoint candidates and then narrows down the keypoint candidates per instance through clustering to estimate individual poses. \textit{HRNet} \cite{sun2019deep} learns high-resolution representations by avoiding direct downsampling in the backbone framework and implementing downsampling through parallel forwarding while maintaining a thread with the original resolution. \textit{AlphaPose} \cite{alphapose, fang2017rmpe} and \textit{YOLOv8-Pose} \cite{yolov8} use top-down approaches to estimate multi-instance poses, first locating the people and then estimating the poses of each individual. \textit{DensePose} \cite{guler2018densepose} estimates the 3D surface through dense pose estimation. At the time we started this work, \textit{YOLOv8-Pose} was the latest pose estimation model. We chose this model to improve the EasyVis framework as it is fast and accurate in inference, fitting our real-time constraints.

\subsection{Dataset Generation}

The dataset generation procedure is typically time-consuming and requires significant labor efforts. In the field of object detection and pose estimation, CORe50 \cite{lomonaco2017core50}, Dex-YCB \cite{chao2021dexycb}, and PoseTrack \cite{andriluka2018posetrack} are fully manually labeled datasets that necessitate a substantial amount of labor to complete. Semi-automated dataset generation methods are employed to reduce these efforts. For instance, T-LESS \cite{hodan2017t} combines manually created 3D models with auto-reconstructed models to decrease the time required to deliver a dataset for object 6D pose estimation. Fully synthetic datasets have been proposed to avoid intensive labor. Examples include the Falling Things (FAT) dataset \cite{tremblay2018falling}, MOTSynth \cite{fabbri2021motsynth}, and SAPIEN \cite{xiang2020sapien}, which feature fully automated annotations. With known physics in virtual environments, numerous datasets can be generated easily. However, creating a high-quality virtual model to approximate real-world objects remains time-consuming, and there are inherent differences between real and virtual environments. On the other hand, our task relies on real-world images to perform point-based 3D reconstruction under a few-view constraint. This reconstruction is sensitive to observation errors, making a high-quality dataset for object pose essential. To maximize efficiency and dataset quality, we decided to use semi-automated methods for dataset generation.

\section{Methodologies}
\subsection{EasyVis Real-time 3D Visualization Framework}

\begin{figure}[!t]
    \centering
    \includegraphics[width=\linewidth]{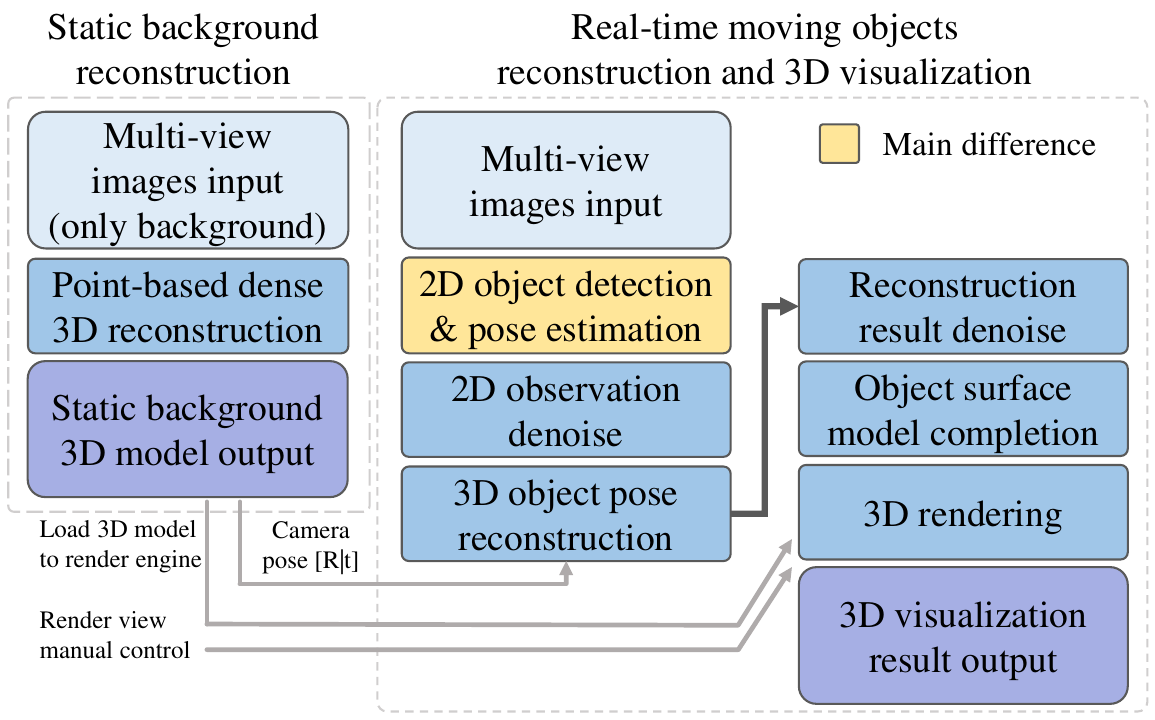}
    \caption{EasyVis 3D reconstruction system workflow. This work is based on this framework, focused on improving 2D object detection and pose estimation module.}
    \label{easyvis}
\end{figure}

EasyVis \cite{sun2025easyvis} is a real-time multi-view rendering system that performs real-time rendering of moving foreground objects while overlaying still (or slowly changing) background scenes. It assumes that foreground objects are rigid and that their 3D models can be obtained offline. During rendering, the 3D positions and orientations (pose) of each object are estimated and tracked by the EasyVis system. Using the known 3D models and estimated poses, each object is then rendered from a viewpoint designated by the user. This approach leads to considerable computational savings compared to re-estimating the 3D models of the foreground objects in each frame. The background is estimated using the traditional structure from motion (SfM) approach \cite{ullman1979interpretation}, with the assumption of a still (or slowly changing) background scene. It can be updated intermittently, significantly reducing the required computation. By combining these two approaches, EasyVis can achieve real-time performance using commercially off-the-shelf computing equipment. The overall workflow of EasyVis is shown in Fig. \ref{easyvis}.


To estimate the poses of foreground objects, EasyVis initially used color markers attached to surgical tools as indicators of 3D keypoints on the objects. While color markers facilitate initial algorithm development, they are not suitable for emulating real-life surgical training operations. Additionally, the effectiveness of color markers can be compromised by changing lighting conditions and confusing background patterns. A key improvement in EasyVis 2 is the use of a deep neural network-based object detection and pose estimation algorithm, YOLOv8-Pose \cite{yolov8}, to detect foreground objects and estimate the 2D poses of each object from individual views.

As shown in Fig. \ref{easyvis}, the main difference of the modified EasyVis framework is in the 2D object detection and pose estimation block. Specifically, EasyVis2 incorporates YOLOv8-Pose for 2D foreground object detection and pose estimation for each view of the multi-camera array. It results in more efficient real-time implementation and better performance. EasyVis2 is implemented in C++, and TensorRT \cite{NVIDIA_TensorRT_2021} is employed to optimize GPU code efficiency.

\begin{figure}[!t]
\centering
\includegraphics[width=\linewidth]{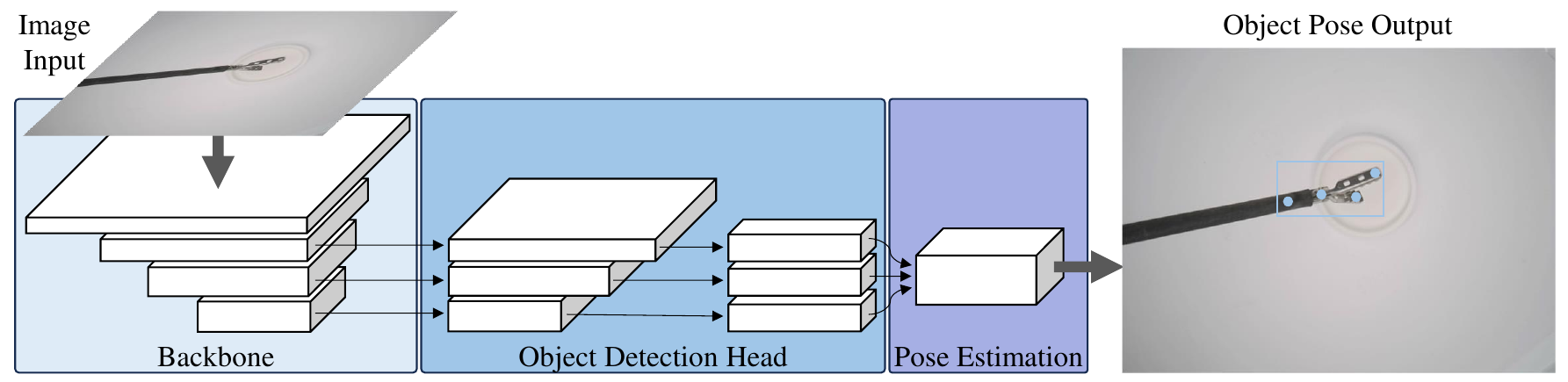}
    \caption{YOLOv8-Pose structure. The neural network first extracts features from the input image in the backbone then detects the object area in the detection head, and then estimates the object pose keypoints in the detected area.}
    \label{yolo}
\vspace{5mm}

\includegraphics[width=\linewidth]{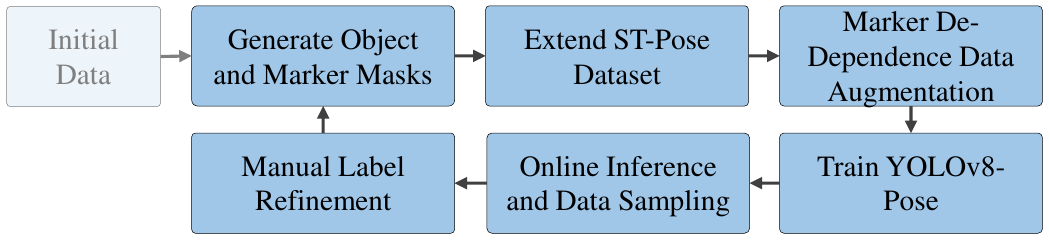}
    \caption{Our efficient training strategy and semi-auto dataset generation pipeline. We train the YOLOv8-Pose model with a gradually extended dataset. The dataset includes the object masks and marker masks generated by the modified Cutie\cite{Cutie}. The marker is used to obtain a consistent surgical tool skeleton keypoint. In data augmentation, we substitute the background and marker with other textures. By doing this, we efficiently achieve a robust model without the dependence on markers.}
    \label{dataGenerate}
\end{figure}

\subsection{YOLOv8-Pose Based 3D Pose Estimation}
\label{3DPose}
YOLOv8-Pose \cite{yolov8} is a deep neural network (DNN) model designed for human pose estimation. The human pose is characterized by a 17-keypoint human pose model with each keypoint corresponding to a body joint position. A COCO human pose dataset \cite{lin2014microsoft} was used to develop the YOLOv8-Pose pre-trained model. 

The deep neural network structure of YOLOv8-Pose is highlighted in Fig. \ref{yolo}. After the input image is presented to the model, the backbone convolutional neural network converts the image into neural network features. Then the object detection head block will estimate the bounding boxes of objects that it has been trained on. The pose estimation block then will estimate the location of keypoints. Finally, both the object detection result (the bounding box, and object class label) and pose estimation result (keypoint locations) are provided as output. 


YOLOv8-Pose was pre-trained using a 17-keypoint human pose model. In EasyVis, we want to estimate the poses of laparoscopic surgical tools which do not share the same pose model of a human body. Thus, the new object needs its own pose model. Referring to Fig. \ref{workflow}(a), the pose model of a laparoscopic surgical grasper consists of four keypoints: two on the tips of two fingers, a third one on the wrist joint, and the fourth one on the arm of the tool. The first three keypoints are selected so that they can be viewed from different cameras in the camera array to facilitate estimation of their 3D positions. The keypoint on the arm, however, is for estimating the 3D orientation of the arm relative to the 3D position of the wrist keypoint. Hence, it is sufficient that the fourth keypoint falls on the center axis of the arm in the image. Note that the function of the cylindrical arm is to deliver the grasper into the training box. Depending on how deep the grasper is inserted into the training box, the length of the arm in the image will vary as shown in Fig. \ref{workflow}(a). This makes it difficult to select a distinct feature point on the arm as the fourth keypoint. Thus, we opt to use the fourth keypoint (together with the third keypoint at the wrist) to mark the center axis of the arm and use the following method to estimate the 3D orientation of the arm relative to the 3D position of the wrist keypoint. 

Let $X_w$ and $X_a$ be respectively the 3D coordinates of the wrist and (a point on the) arm keypoint of the grasper, and $O_i$ be the 3D coordinate of the optical center of the $i^{th}$ camera in the camera array. These three points determine a plane in the 3D space. The normal vector of this plane can be found from the cross product of two vectors: Denote $\overrightarrow{O_i X_w}$ and $\overrightarrow{O_i X_a}$ to be vectors from $O_i$ to $X_w$ and $X_a$ respectively. The normal vector can be found as
\begin{equation}
\label{eq1}
    n_i=\overrightarrow{O_i X_w} \times \overrightarrow{O_i X_a}
\end{equation}
Define a matrix $N$ consisting of such normalized vectors from each view of the multi-view camera array, ${\bf N} =[n_1\ n_2\ ...\ n_I]^T$. Denote $u$ to be a normalized 3D vector from $X_a$ to $X_w$, namely the 3D orientation of the arm of the grasper. Then we have:
\begin{equation}
\label{eq2}
    {\bf N} u=0
\end{equation}
$u$ may be estimated as the right singular vector corresponding to the minimum singular value of the matrix ${\bf N}$, as long as the number of views used $I > 3$.

To estimate the 3D coordinate of the wrist keypoint  $X_w$ as well as those of the two fingertips of the grasper from corresponding 2D coordinates estimated using Yolov8-Pose, triangulation may be applied: Denote $(x_{wi},y_{wi})$ to be the estimated 2D coordinates of the wrist keypoint in the $i^{th}$ view, and $\mathbf{M}_i = \left[ \begin{matrix} \mathbf{m}_{1,i}^T & \mathbf{m}_{2,i}^T & \mathbf{m}_{3,i}^T  \end{matrix}  \right]^T$ to be the corresponding $3 \times 4$ camera projection matrix, then according to the pin-hole camera equation, one may write

\begin{equation}
    \left[\begin{matrix}1 & 0 & -x_{wi} \\ 0 & 1 & -y_{wi} \end{matrix}\right] \left[ \begin{matrix}\mathbf{m}_{1,i} \\ \mathbf{m}_{2,i} \\ \mathbf{m}_{3,i}  \end{matrix} \right]X_w = \mathbf{H}_i \mathbf{M}_i X_w = \mathbf{0}
\end{equation}
where
\begin{equation*}
    \mathbf{H}_i = \begin{bmatrix}
        1 & 0 &  -x_{wi}\\
        0 & 1 & -y_{wi}\\
    \end{bmatrix}
\end{equation*}
Then, combining all $I$ views, one has
\begin{equation}
    \begin{bmatrix}
        \mathbf{H}_1 \mathbf{M}_1\\
        \vdots\\
        \mathbf{H}_I \mathbf{M}_I\\
    \end{bmatrix}X_w = \mathbf{G}X_w=0
\end{equation}
Since the $\mathbf{H}_i$ depends on the estimated 2D keypoint coordinates, and $\mathbf{M}_i$ is obtained from the camera extrinsic parameters, the $\mathbf{G}$ matrix is readily available. Then, $X_w$ may be estimated as the minimum singular value of the $\mathbf{G}$ matrix. The same procedure may be used to estimate the 3D coordinates of the two fingers.

\subsection{ST-Pose Dataset}

Since the Yolov8-Pose pre-trained model was developed for human body pose estimation, it needs to be modified to detect and estimate the pose of laparoscopic surgical tools. Additionally, we need to develop a surgical tool training dataset (ST-Pose) to fine-tune the Yolov8-Pose pre-trained model for use in EasyVis2.


\textbf{Dataset Description}. ST-Pose is designed around the laparoscopic surgical training box bean drop task. Images from videos captured by cameras of the EasyVis2 camera array will be used as input to train the Yolov8-Pose model. For each training image, manual labeling of the bounding boxes and keypoints of surgical tools and beans were needed. Hence, labeling is a labor-intensive task.


The definition of pose (keypoints) for each surgical tool must be chosen so that it is associated with a distinct shape feature that is invariant to multiple views. An example of the selection of keypoints for a surgical grasper has been discussed in section \ref{3DPose}. Specifically, keypoints are selected on the tips of two fingers and on the wrist of the grasper because these positions are unique and invariant to different views. Due to symmetry, there is no need to distinguish the two fingers. 

The keypoint in the cylindrical arm of the grasper will be treated differently because there is no unique position on the arm that can be easily distinguished and detected by multiple cameras. To make the labeling task easier, a visible marker was placed on the arm so that a keypoint may be placed consistently by the label annotator. This visible marker is removed in the image when it is used for training the Yolov8-Pose model. This arm marker needs to be placed along the axis of the cylindrical arm so that a line from the wrist keypoint to the arm keypoint defines the orientation of the arm. Procedures using the arm keypoint to estimate the arm orientation are discussed in Eq. (\ref{eq1}) and eq. (\ref{eq2}). 

\begin{figure}[!t]
\centering
\includegraphics[width=\linewidth]{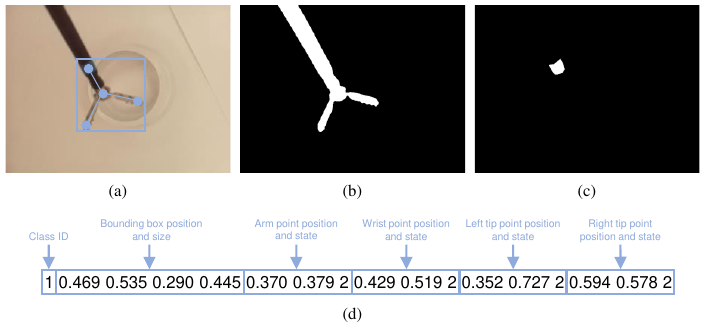}
    \caption{One set of samples in the ST-Pose dataset. The captured image simulates the view under a laparoscope, with only the functional head visible. (a) A surgical grasper with a bounding box and object pose is defined by a box and four keypoints. (b) Object mask covering the surgical tool area. (c) Marker mask covering the marker area. (d) The label that describes the object pose states.}
    \label{stpose}
    \end{figure}

Fig. \ref{stpose} (a) shows the captured image with the annotated bounding box and keypoints. During annotation, a surgical tool mask shown in Fig. \ref{stpose} (b) is produced automatically using a modified segmentation algorithm \cite{Cutie} to facilitate the annotator to identify the silhouette of the object. The location of the visible marker mask is shown in Fig. \ref{stpose} (c). This marker is removed during the model training. An example of the text label of a training sample is shown in (d). Masks were generated using a semi-auto segmentation algorithm Cutie \cite{Cutie}.

This ST-Pose dataset includes 2,930 images with 3,682 labeled grasper instances and 3,198 labeled bean instances. We provided a separately-sampled validation set with 300 images. Additionally, we introduced an LS scissor into our dataset as an extension. It's separate from the LS grasper. This extension includes 422 images with 422 labeled instances. The experiments in this paper only focus on the LS grasper part. The image size is 640 × 480 pixels. In total, there are 3,652 images, 3,632 object masks, and 2,804 marker masks. The idea of LS pose can be generalized and applied to surgical tools with similar structures, such as graspers of various sizes.

\begin{figure}[!t]
    \centering
    \includegraphics[width=\linewidth]{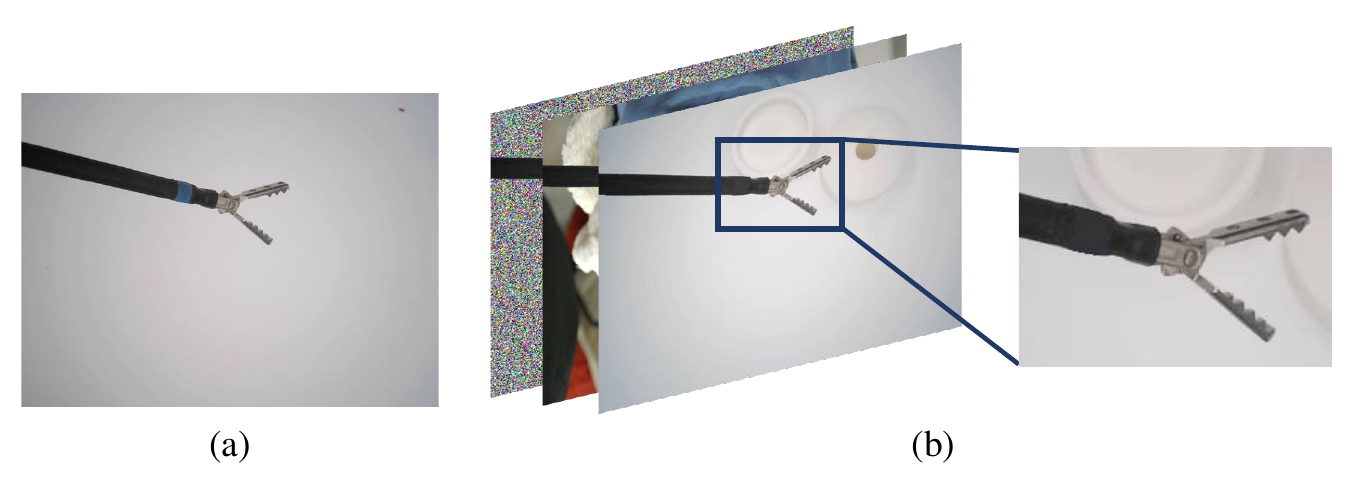}
    \caption{Demo of data augmentation using masks: a) Original image. b) Data augmentation results. The background is substituted using different random textures and the marker is substituted with rod texture. The model trained with the augmented dataset can avoid treating the background as part of the object and avoid estimating keypoints relying on the marker.}
    \label{dataaug}
\end{figure}


\begin{figure*}[!t]
    \centering
    \includegraphics[width=\linewidth]{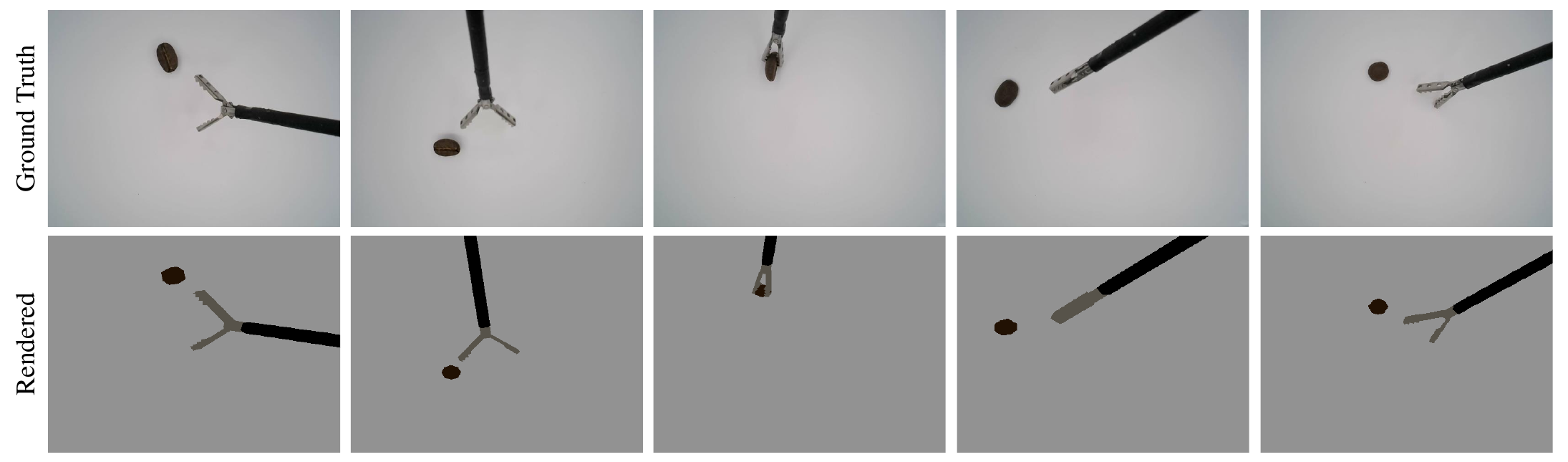}
    \caption{We back-projected the reconstructed 3D scene to the 2D ground truth view. Our goal is to approximate the reconstructed scene to the real scene. Back-projecting the reconstructed scene and comparing it to the ground truth 2D image indicates the precision of our system.}
    \label{visualization}
\end{figure*}

\textbf{Dataset Creation Method}
To create the ST-Pose dataset, we modify the EasyVis2 processing pipeline so that the 2D object detection and pose estimation outcome from cameras on the camera arrays will be output to a video screen. An operator will monitor this screen and mark images (video frames) whose outcome is erroneous. A commercial tool Roboflow (https://roboflow.com/annotate) is used to annotate these marked images to correct the annotation error. We develop an experiment procedure that includes a set of tasks that are likely to occur during EasyVis2 training sessions. With These, the dataset creation is an iterative process which is summarized as follows:

{\it Initial Training}: The generic Yolov8-Pose pre-trained model will be trained with a few (around 40) images of surgical tools to bootstrap its capability to detect and estimate the pose of a foreground object used in EasyVis.

{\it Iteration}: In each iteration,
\begin{itemize}
    \item[a] The most recently updated Yolov8-Pose model will be used in an experiment using the modified pipeline to predict the location and pose of foreground objects.
    \item[b]  A human operator will monitor the outcome and mark video frames exhibiting location and pose estimation error.
    \item[c]  A human annotator then will use Roboflow to provide correct labeling of these images. These images will be incorporated into the training dataset. 
    \item[d] Then, Yolov8-Pose will be fine-tuned with these newly added training samples. 
\end{itemize}

The above development method ST-Pose dataset has several unique features: First, it integrates model improvement and dataset development into an integrated iterative process. The model is developed once the dataset is created. Next, it employed a greedy sample selection strategy such that only samples mistakenly predicted by the current model will be labeled.

\section{Experiments}
In this section, we report four experiments to evaluate the performance of the EasyVis2 system presented in this work. The first experiment provides a qualitative validation of a simulated laparoscopic surgical task using EasyVis2. The second experiment evaluates the processing time of EasyVis2. The third experiment evaluates the accuracy of EasyVis2 3D rendering results. The last experiment focuses on the performance of Yolov8-Pose alone under different dataset setups.

\subsection{3D Rendering Examples}



To verify that the 3D rendering provides the correct scene, we back-projected the reconstructed 3D model to viewpoints that coincide with the cameras on the camera array and compared the back-projected image with the image taken by the camera. A perfect match indicates accurate 3D model rendering. In Fig. \ref{visualization}, the images captured by five cameras are shown on the upper row. The back projections of the reconstructed 3D model toward these five cameras are shown in the bottom row. It is quite clear that there is a good match between the images of these two rows.

\begin{figure}[!t]
    \centering
    \includegraphics[width=\linewidth]{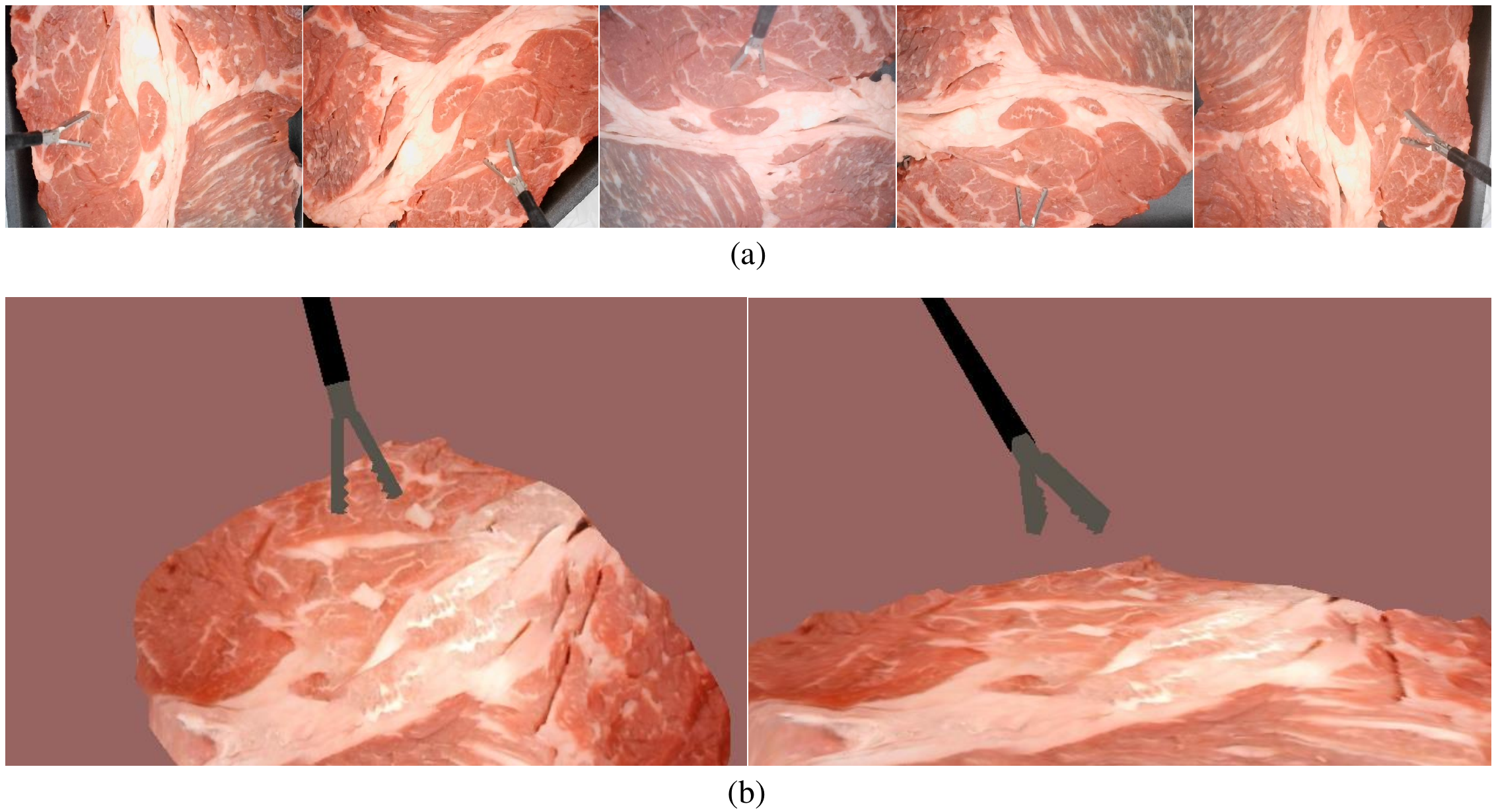}
    \caption{3D rendering results on animal tissue. a) Five observation views from the camera array. b) Render results with a surgical tool. Our work provides an additional sense of depth through rendering the side view. A distance between surgical tools and the tissue is visible from the side view.}
    \label{realtissue}
    \vspace{5mm}
\end{figure}

In Fig. \ref{realtissue}, we show the 3D rendering results using real-world animal tissue (beef steak). Five camera views are shown in Fig. \ref{realtissue}(a). Structure from motion (SfM) \cite{ullman1979interpretation} and multiview stereo (MVS) \cite{furukawa2015multi} algorithms are used to capture and render the 3D model of the animal tissue background. 
In Fig. \ref{realtissue}(b), two projected scenes are shown. These are projected sideways so that the distance between the grasper and the surgical surface can be seen.

\subsection{Timing Analysis}

\begin{figure}[!t]
    \centering    
    \includegraphics[width=\linewidth]{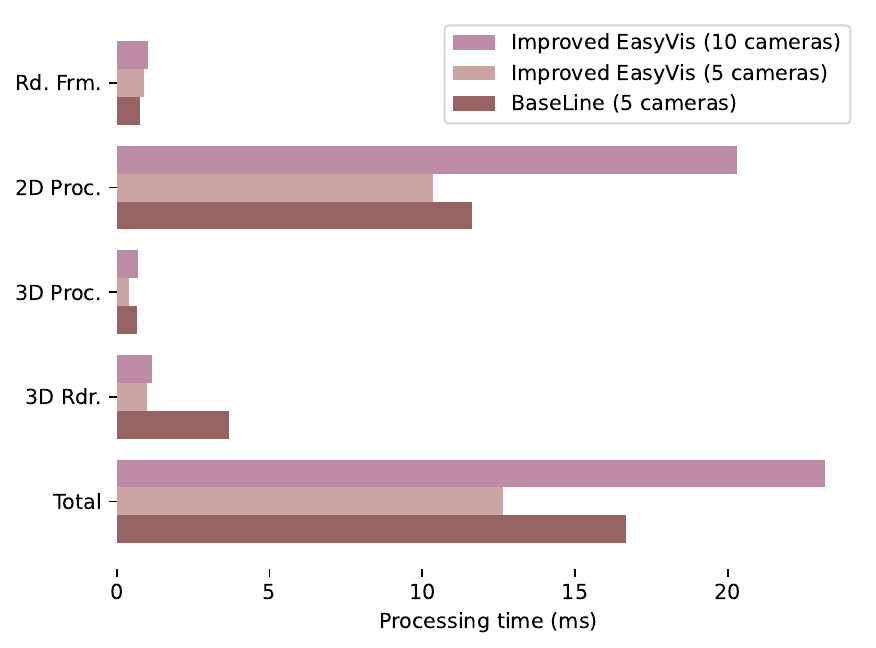}
    \caption{Processing time comparison. We compared the processing times for the baseline with 5 cameras, the EasyVis2 with 5 cameras, and the  EasyVis2 with 10 cameras.}
    \label{timeResult}
\end{figure}

In the second experiment, the processing time (second per frame) under different setup conditions is compared against that of the baseline algorithm (EasyVis). The results are summarized in Fig. \ref{timeResult}. The processing pipeline is divided into four stages: (a) reading a new frame from all cameras of the camera array. Multi-thread parallel computing is used here to speed up computing. (b) 2D object detection and pose estimation using adapted Yolov8-Pose. This process runs in the GPU sequentially one view after another. (c) 3D reconstruction and 3D object pose estimation.  Finally, (d) the estimated 3D models of foreground objects and background scenes are sent to GPU's rendering engine to be projected to desired viewpoints. 

The first timing comparison results are shown in Fig. \ref{timeResult}. We used five cameras (1 camera array) and ten cameras (2 camera arrays) in EasyVis2 and only used five cameras (1 camera array) in EasyVis which was designed only for five cameras. From this figure, one notes that EasyVis2 takes a bit longer to read view frames than EasyVis. This is because different cameras are used in EasyVis2. Obviously, the multiview 2D object detection and pose estimation stage dominate the computing time. Yet using Yolov8-Pose does reduce the computing time (about 1.3ms per frame) compared to that of EasyVis which uses a different 2D object detection and pose estimation method. Since it is sequentially processed in the GPU, ten cameras take twice as long as five cameras. We also noticed that the 3D rendering and projection (stage d) take significantly less time compared to that of EasyVis. 

Overall, the 5-camera configuration achieves a total frame processing time of 12.6 ms, which is 4 ms and 24.2\% faster than EasyVis. Even though the total processing time is longer with the 10-camera setup, the system still achieves real-time performance, with a time cost of 23.2 ms per frame.

\begin{table}[!t]
    \caption{3D reconstruction evaluation using BPEs. We compare the EasyVis-V2 to baseline EasyVis-V1 and the accuracy under different setups. Lower is better.}
  \centering
  \begin{tabular}{l|l|ccc}
    \hline
     & & Bean & Grasper & Avg.\\ 
    \hline
    \multirow{3}{*}{\parbox{1.4cm}{baseline \\ (5 views)}} & BPE\_PD & 18.025 & 7.336&  12.681\\
     & BPE\_PPw & 2.816\% &  1.146\% &1.981\%\\
     & BPE\_PPh & 3.755\% & 1.528\% & 2.642\%\\
     \hline
     \multirow{3}{*}{\parbox{1.4cm}{EasyVis-V2 \\ (10 views)}} & BPE\_PD & 3.761& 6.607& 5.184\\
     & BPE\_PPw & 0.588\% & 1.032\% &  0.810\%\\
     & BPE\_PPh & 0.784\% & 1.376\% & 1.080\%  \\
     \hline
     \multirow{3}{*}{\parbox{1.4cm}{EasyVis-V2 \\ (9 views)}} & BPE\_PD & 3.863& 6.871& 5.367\\
     & BPE\_PPw & 0.604\% & 1.074\% &  0.839\%\\
     & BPE\_PPh & 0.805\% & 1.431\% & 1.118\%  \\
     \hline
     \multirow{3}{*}{\parbox{1.4cm}{EasyVis-V2 \\ (8 views)}} & BPE\_PD & 3.419& 7.210& 5.315\\
     & BPE\_PPw & 0.534\% & 1.127\% &  0.831\%\\
     & BPE\_PPh & 0.712\% & 1.502\% & 1.107\%  \\
     \hline
     \multirow{3}{*}{\parbox{1.4cm}{EasyVis-V2 \\ (7 views)}} & BPE\_PD & 2.851& 5.620& 4.236\\
     & BPE\_PPw & 0.445\% & 0.878\% &  0.662\%\\
     & BPE\_PPh & 0.594\% & 1.171\% & 0.884\%  \\
     \hline
     \multirow{3}{*}{\parbox{1.4cm}{EasyVis-V2 \\ (6 views)}} & BPE\_PD & 3.248& 5.782& 4.515\\
     & BPE\_PPw & 0.508\% & 0.903\% &  0.706\%\\
     & BPE\_PPh & 0.677\% & 1.205\% & 0.941\%  \\
     \hline
     \multirow{3}{*}{\parbox{1.4cm}{EasyVis-V2 \\ (5 views)}} & BPE\_PD & 3.568& 6.738& 5.153\\
     & BPE\_PPw & 0.558\% & 1.053\% &  0.806\%\\
     & BPE\_PPh & 0.743\% & 1.404\% & 1.074\%  \\
     \hline
     \multirow{3}{*}{\parbox{1.4cm}{EasyVis-V2 \\ (4 views)}} & BPE\_PD & 3.561& 6.752& 5.157\\
     & BPE\_PPw & 0.556\% & 1.055\% &  0.806\%\\
     & BPE\_PPh & 0.742\% & 1.407\% & 1.075\%  \\
     \hline
     \multirow{3}{*}{\parbox{1.4cm}{EasyVis-V2 \\ (3 views)}} & BPE\_PD & 3.020& 7.974& 5.497\\
     & BPE\_PPw & 0.472\% & 1.246\% &  0.859\%\\
     & BPE\_PPh & 0.629\% & 1.661\% & 1.145\%  \\
     \hline
     \multirow{3}{*}{\parbox{1.4cm}{EasyVis-V2 \\ (2 views)}} & BPE\_PD & \textbf{2.210}& \textbf{5.407}& \textbf{3.809}\\
     & BPE\_PPw & 0.345\% & 0.845\% &  0.595\%\\
     & BPE\_PPh & 0.460\% & 1.126\% & 0.793\%  \\

    \hline
  \end{tabular}
  \label{tab:bpe}
\end{table}

\begin{table}[!t]
    \caption{Precision study for PE under different models and validation sets. The models are trained with different kinds of data augmentation methods.}
  \centering
  \begin{tabular}{c|c|cccc}
    \hline
    \makecell{Tasks} & \makecell{Matrices} & \makecell{Val. $S_1$} & \makecell{Val. $S_2$} & \makecell{Val. $S_3$} & \makecell{Val. $S_4$} \\
    \hline
    \multicolumn{6}{l}{Mdl. $M_1$} \\
    \hline
    \multirow{4}{*}{\makecell{OD}}   & Precision& 0.993 & 0.996 & 0.580 & 0.587\\
                                                     & Recall& 0.993 & 0.987 & 0.342 & 0.320\\
                                                     & mAP@50 & 0.995 & 0.995 & 0.335 & 0.317\\
                                                     & mAP@50:95 & 0.954 & 0.856 & 0.278 & 0.264\\
    \hline
    \multirow{4}{*}{\makecell{PE}}   & Precision& 0.993 & 0.991 & 0.580 & 0.592\\
                                                     & Recall& 0.993 & 0.994 & 0.347 & 0.325\\
                                                     & mAP@50 & 0.995 & 0.995 & 0.343 & 0.326\\
                                                     & mAP@50:95 & 0.994 & 0.994 & 0.328 & 0.313\\

    \hline
    \multicolumn{6}{l}{Mdl. $M_2$} \\
    \hline
    \multirow{4}{*}{\makecell{OD}}   & Precision& 0.998 & 0.998 & 0.668 & 0.714\\
                                                     & Recall& 0.994 & 0.994 & 0.288 & 0.306\\
                                                     & mAP@50 & 0.995 & 0.995 & 0.308 & 0.335\\
                                                     & mAP@50:95 & 0.945 & 0.958 & 0.262 & 0.285\\
    \hline
    \multirow{4}{*}{\makecell{PE}}   & Precision& 0.998 & 0.997 & 0.653 & 0.716\\
                                                     & Recall& 0.994 & 0.993 & 0.293 & 0.309\\
                                                     & mAP@50 & 0.995 & 0.995 & 0.312 & 0.339\\
                                                     & mAP@50:95 & 0.995 & 0.995 & 0.305 & 0.332\\
                                                     
    \hline
    \multicolumn{6}{l}{Mdl. $M_3$} \\
    \hline
    \multirow{4}{*}{\makecell{OD}}   & Precision& 0.977 & 0.997 & 0.943 & 0.942\\
                                                     & Recall& 0.995 & 0.995 & 0.977 & 0.977\\
                                                     & mAP@50 & 0.995 & 0.995 & 0.986 & 0.985\\
                                                     & mAP@50:95 & 0.959 & 0.930 & 0.941 & 0.938\\
    \hline
    \multirow{4}{*}{\makecell{PE}}   & Precision& 0.989 & 0.989 & 0.936 & 0.935\\
                                                     & Recall& 0.987 & 0.988 & 0.970 & 0.969\\
                                                     & mAP@50 & 0.988 & 0.986 & 0.976 & 0.975\\
                                                     & mAP@50:95 & 0.910 & 0.903 & 0.901 & 0.902\\

    \hline
    \multicolumn{6}{l}{Mdl. $M_4$} \\
    \hline
    \multirow{4}{*}{\makecell{OD}}   & Precision& 0.995 & 0.995 & 0.956 & 0.967\\
                                                     & Recall& 0.997 & 0.997 & 0.961 & 0.960\\
                                                     & mAP@50 & 0.995 & 0.995 & 0.987 & 0.989\\
                                                     & mAP@50:95 & 0.972 & 0.970 & 0.957 & 0.959\\
    \hline
    \multirow{4}{*}{\makecell{PE}}   & Precision& 0.994 & 0.994 & 0.955 & 0.966\\
                                                     & Recall& 0.996 & 0.996 & 0.960 & 0.959\\
                                                     & mAP@50 & 0.995 & 0.995 & 0.986 & 0.987\\
                                                     & mAP@50:95 & 0.995 & 0.995 & 0.985 & 0.987\\




    \hline
  \end{tabular}
  \label{modelVsVal}
\end{table}

\subsection{3D Reconstruction Evaluation}
\begin{figure}[!t]
    \centering
    \includegraphics[width=\linewidth]{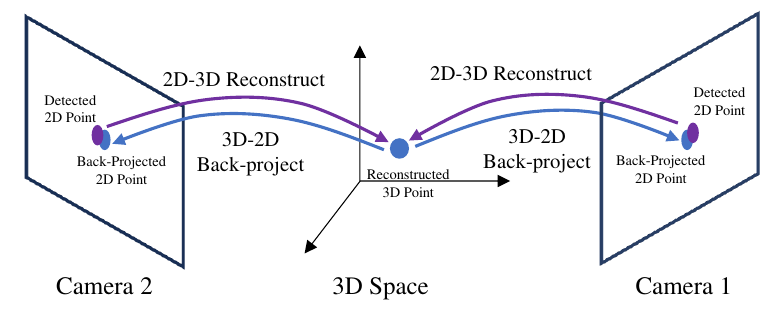}
    \caption{Concept of back-project error. Consider two cameras, each observing a point in 3D space. With two 2D observed points from the two cameras, the corresponding 3D point can be estimated. Next, the estimated 3D point is back-projected onto the original camera views in the image coordinate. The back-projected point is then compared to the detected point, and the distance between two 2D points is calculated to determine the back-projection error.}
    \label{bpe}
\end{figure}

In this experiment, we used Back Project Error (BPE) \cite{reference} as the performance metric to evaluate the 3D reconstruction quality of the improved system. BPE back-projects the reconstructed 3D points to 2D and then compares the distance from these points to the original detected points, testing the accuracy of 3D reconstruction. The concept of BPE is shown in Fig. \ref{bpe}. We back-project the reconstructed points to the original camera views and calculate the distance between the original detected points and back-project points.

Suppose we have a 3D point (X) obtained from 3D reconstruction, then we back-project it from 3D space to an image:
\begin{equation}
\tilde{x}=K[R|t]X    
\end{equation}
where $\tilde{x}$ is the 2D coordinate in the image, $K$ is the camera intrinsic matrix, $[R|t]$ is the camera extrinsic matrix. We then can calculate an Euclidean distance from this point to the predicted 2D point $\hat{x}$ in pixel space, one of the 2D points to reconstruct the 3D point $X$. Considering all reconstructed points in all camera views, we have BPE\_PD:

\begin{equation}
\it{BPE}_{\_PD} = \frac{1}{N} \sum_{i} \|({\tilde{x}_{i}} - {\hat{x}}_{i})\|^2_2 
\end{equation}
where $N$ is the total number of back-projected points in all camera views and $i=0,1,...,N$.

Moreover, we used two additional representations to describe the BPE: BPE in pixel proportion in image width $BPE_{\_PPw}$ and BPE in pixel proportion in image height $BPE_{\_PPh}$. These two representations help to evaluate the error level objectively on the whole image scale. Lower values indicate better keypoint detection consistency among different views and various object poses, resulting in a lower reconstruction error. Given two camera arrays with five cameras on each array, we have ten cameras in total. In this experiment, for each number of camera views, we take the average result of all the combinations. We summarized the results in Table \ref{tab:bpe}.

Comparing EasyVis-V2 to EasyVis-V1 under the 5-camera setup, EasyVis-V2 has higher accuracy. It achieves an average BPE\_PD of 5.153, BPE\_PPw of 0.806\%, and BPE\_PPh of 1.074\%, which represents a 59.4\% improvement compared to the baseline. Still based on this setup, EasyVis-V2 achieves a BPE\_PD of 3.568 for the bean and 6.752 for the grasper, which represents 80.2\% and 8.0\% improvements compared to the baseline. Additionally, when conducting ablation studies on EasyVis-V2 using different numbers of cameras, we can see that the results of using 2-10 cameras all perform better than the baseline. The BPEs from the 2-camera to 10-camera setups show fluctuations due to the stochastic properties of the testing sample under distortion of the camera or partial occlusion of the objects. The two-view result shows the lowest error. This is because with only two views, the error rate is minimized, as it only contains errors from the camera calibration (SfM). In contrast, using more than three views introduces the least squares property, which averages the observation noise and can result in a higher error rate.

Benefiting from the DNN-based pose estimation, the detected object keypoints are less sensitive to object shape and environmental light, resulting in improved system robustness. Additionally, due to fewer post-processing procedures to handle edge cases using information from previous frames, the system is highly responsive to object movement. This responsiveness results in the reconstructed 3D poses being more similar to the ground truth when the grasper is moving, even though the previous results are used to denoise, which is reflected in the better BPEs compared to the baseline.

\begin{figure*}[!t]
    \centering
    \includegraphics[width=\linewidth]{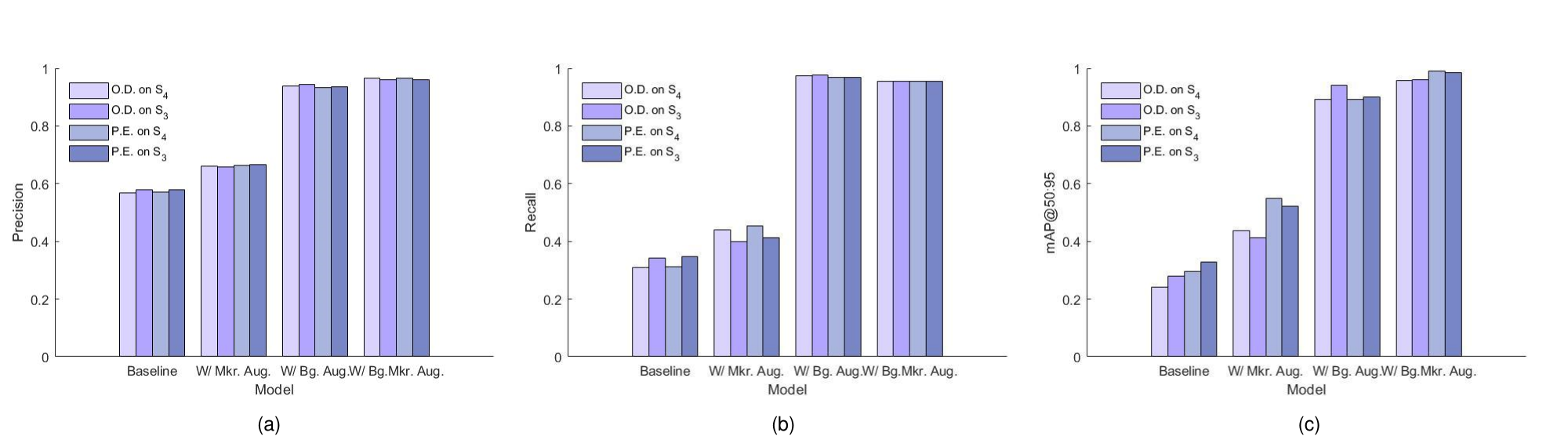}
    \caption{Performance of models trained with different data augmentation setup on the validation sets with background augmentation $S_3$ and with background and marker augmentation $S_4$.}
    \label{performanceAug1}
    \vspace{5mm}
    \includegraphics[width=\linewidth]{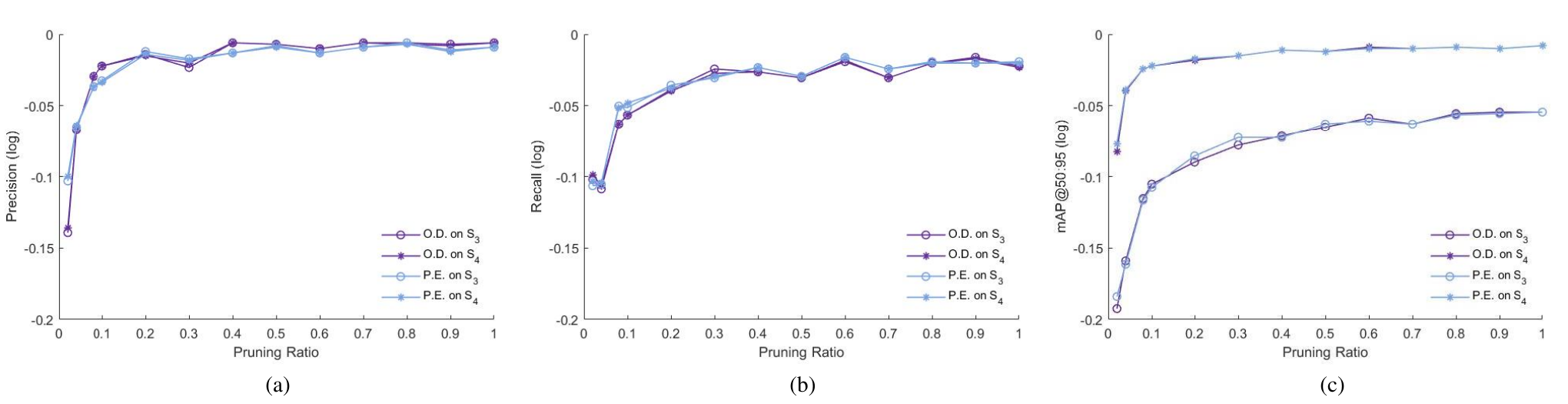}
    \caption{The performance of models trained with the pruned dataset was evaluated on the validation sets $S_3$ and $S_4$. The performance metrics include accuracy, precision, and recall.}
    \label{performanceAug2}
\end{figure*}

\begin{table*}[!t]
    \caption{Performance of models trained with the pruned dataset on the validation set $S_3$.}
  \centering
  \begin{tabular}{c|c|cccccccccccccc}
    \hline
    Tasks & Matrices  & \makecell{$2\%$} & \makecell{$4\%$} & \makecell{$6\%$}  & \makecell{$8\%$}  & \makecell{$10\%$}  & \makecell{$20\%$}  & \makecell{$30\%$}  & \makecell{$40\%$}  & \makecell{$50\%$}  & \makecell{$60\%$}  & \makecell{$70\%$} & \makecell{$80\%$} & \makecell{$90\%$} & \makecell{$100\%$} \\
    \hline
    \multirow{4}{*}{\makecell{OD}}   & Precision                 & 0.885 & 0.926 & 0.970 & 0.978& 0.975& 0.989& 0.984& 0.993& 0.994& 0.985& \textbf{0.995}& 0.988& 0.989& \textbf{0.995}\\
                                                     & Recall    & 0.860 & 0.950 & 0.933 & 0.958& 0.952& 0.963& 0.962& 0.974& 0.966& 0.966& 0.971& \textbf{0.985}& 0.973& 0.977\\
                                                     & mAP@50    & 0.940 & 0.972 & 0.979 & 0.985& 0.986& 0.989& 0.989& 0.992& 0.991& 0.991& 0.992& 0.992& 0.991& \textbf{0.993}\\
                                                     & mAP@50:95 & 0.825 & 0.885 & 0.892 & 0.907& 0.906& 0.922& 0.927& 0.940& 0.936& 0.942& 0.946& 0.946& \textbf{0.948}& 0.947\\
    \hline
    \multirow{4}{*}{\makecell{PE}}   & Precision                 & 0.882 & 0.926 & 0.970 & 0.978& 0.972& 0.989& 0.984& 0.993& 0.994& 0.985& \textbf{0.995}& 0.988& 0.989& \textbf{0.995}\\
                                                     & Recall    & 0.867 & 0.950 & 0.933 & 0.958& 0.956& 0.963& 0.962& 0.973& 0.966& 0.966& 0.971& \textbf{0.985}& 0.972& 0.976\\
                                                     & mAP@50    & 0.942 & 0.972 & 0.979 & 0.984& 0.985& 0.988& 0.988& 0.991& 0.991& 0.990& \textbf{0.992}& \textbf{0.992}& 0.990& \textbf{0.992}\\
                                                     & mAP@50:95 & 0.926 & 0.967 & 0.975 & 0.980& 0.981& 0.986& 0.986& 0.989& 0.989& 0.988& 0.990& 0.990& 0.989& \textbf{0.991}\\                                  
                                 
    \hline
  \end{tabular}
  \label{prungTestS3}
\end{table*}

\begin{table*}[!t]
    \caption{Performance of models trained with the pruned dataset on the validation set $S_4$.}
  \centering
  \begin{tabular}{c|c|cccccccccccccc}
    \hline
    Tasks & Matrices  & \makecell{$2\%$} & \makecell{$4\%$}  & \makecell{$6\%$} &\makecell{$8\%$}  & \makecell{$10\%$}  & \makecell{$20\%$}  & \makecell{$30\%$}  & \makecell{$40\%$}  & \makecell{$50\%$}  & \makecell{$60\%$}  & \makecell{$70\%$} & \makecell{$80\%$} & \makecell{$90\%$} & \makecell{$100\%$} \\
    \hline
    \multirow{4}{*}{\makecell{OD}}   & Precision                 & 0.890 & 0.927 & 0.969 & 0.980& 0.970& 0.986& 0.984& \textbf{0.991}& 0.985& 0.983& 0.986& \textbf{0.991}& 0.985& \textbf{0.991}\\
                                                     & Recall    & 0.869 & 0.950 & 0.936 & 0.960& 0.958& 0.967& 0.971& 0.976& 0.973& 0.969& 0.980& \textbf{0.986}& 0.975& 0.981\\
                                                     & mAP@50    & 0.944 & 0.973 & 0.979 & 0.987& 0.985& 0.990& 0.990& 0.993& 0.992& 0.991& 0.993& 0.993& 0.992& \textbf{0.994}\\
                                                     & mAP@50:95 & 0.833 & 0.887 & 0.894 & 0.910& 0.908& 0.924& 0.931& 0.939& 0.940& 0.942& 0.946& 0.948& 0.947& \textbf{0.949}\\
    \hline
    \multirow{4}{*}{\makecell{PE}}   & Precision                 & 0.890 & 0.927 & 0.970 & 0.979& 0.970& 0.986& 0.983& 0.991& 0.985& 0.983& 0.985& 0.991& 0.985& \textbf{0.992}\\
                                                     & Recall    & 0.869 & 0.950 & 0.936 & 0.960& 0.958& 0.967& 0.970& 0.975& 0.973& 0.969& 0.979& \textbf{0.986}& 0.975& 0.980\\
                                                     & mAP@50    & 0.944 & 0.973 & 0.980 & 0.986& 0.985& 0.989& 0.989& 0.991& 0.991& 0.990& \textbf{0.992}& \textbf{0.992}& 0.991& \textbf{0.992}\\
                                                     & mAP@50:95 & 0.929 & 0.967 & 0.977 & 0.982& 0.981& 0.987& 0.987& 0.990& 0.990& 0.988& 0.990& \textbf{0.991}& 0.989& \textbf{0.991}\\                           
                                 
    \hline
  \end{tabular}
  \label{prungTestS4}
\end{table*}

\subsection{Yolov8-Pose Pose Estimation Accuracy}
In this test, we conducted ablation studies for the 2D processing module to compare the performance of different models on various validation sets, using precision and recall. We evaluated the performance of pose estimation and object detection. Object detection serves as an intermediate process, and we include it in our evaluation since it is part of our dataset, formatted according to the standard COCO keypoint dataset. Precision refers to the proportion of correctly detected objects and estimated poses relative to the total number of bounding boxes or poses predicted by the model, while recall describes the proportion of correct predictions relative to the total number of labels. The correctness is based on Intersection over Union (IoU) and Intersection over Union (IoU), if the value is greater than a given threshold, the prediction is treated as correct. A model with higher precision and recall scores indicates better performance. This experiment helps us better understand the relationship between model performance and training data diversity.


IoU is used to evaluate object detection performance. This metric uses the ratio of the intersection of the prediction and ground truth to their union, as follows: 
\begin{equation}
    IoU=\frac{A_{pred}\cap A_{GT}}{A_{pred}\cup A_{GT}}
\end{equation}
where $A_{pred}$ is the predicted bounding box area and $A_{GT}$ is the ground truth bounding box area. OKS is used to evaluate pose estimation. It describes how close a predicted point is to the ground truth, as follows:
\begin{equation}
OKS = \frac{\sum_i \exp \left( -\frac{d_i^2}{2s^2 k_i^2} \right)\mathds{1}(v_i > 0)}
{\sum_i \mathds{1}(v_i > 0)}
\end{equation}
where $d_i$ is the Euclidean distance of the $i^{th}$ predicted point to the ground truth, $s = \sqrt{W \times H}$ is the object scale defined by the square root of the product of the detected object's bounding box width and height, $k_i$ is the keypoint-specific constant as a hyperparameter, and $v_i$ is the visibility score. If the score is 0, the point is invisible and skipped in the OKS calculation. Additionally, mAP@50 and mAP@50:95 are used to evaluate the model. mAP@50 calculates the mean average precision with an acceptance threshold of 0.5, which helps evaluate the model's basic detection ability. mAP@50:95 calculates the mean average precision with thresholds from 0.5 to 0.95 in 0.05 intervals, which helps evaluate the object or keypoint localization ability. Higher values indicate better performance.

The test models are based on the YOLOv8-Pose-m model and trained with ST-Pose under different data augmentation conditions. These models utilize pre-trained weights from the COCO dataset, which enhances their feature extraction capabilities. The data augmentations include no augmentation ($M_1$), marker substitution ($M_2$), background substitution ($M_3$), and both background and marker substitution ($M_4$). Correspondingly, there are four validation sets using these augmentations: validation set without augmentation ($S_1$), with marker substitution ($S_2$), with background substitution ($S_3$), and with both background and marker substitution ($S_4$). The augmented validation sets effectively reflect the performance of the trained models in more complex environments.

We summarized the results in Table \ref{modelVsVal} and Fig. \ref{performanceAug1}. These results include the performance in the tasks of pose estimation (PE) and object detection (OD) as byproducts of the YOLOv8-Pose model. The results indicate that the model trained with a background-augmented dataset reduces the overfitting problem introduced by the singular background in the laparoscopic training box environment. Comparing $M_1$ to $M_3$ in PE, the precision increases from 0.580 to 0.936 on $S_3$, representing a 61.4\% performance improvement. Additionally, applying marker removal augmentation enhances precision performance. Comparing the PE of model $M_3$ with $M_4$, the precision increases from 0.936 to 0.955 in $S_3$, and from 0.935 to 0.966 in $S_4$, which are improvements of 2.0\% and 3.2\%, respectively. Models trained with marker substitution demonstrate a better ability to estimate keypoints by perceiving a larger image area instead of relying on marker patterns. The model with both background and marker substitution achieves higher precision without relying on the marker. The data augmentations improve the training data diversity, resulting in models with better performance in more complicated environments. Consequently, our method has a higher chance of working in real surgery, even though we haven't trained the model with real surgery images, which avoids the difficulty of sampling real surgery data. Moreover, data augmentations help us generate a strong dataset efficiently under our simple box trainer setup.


In addition, we conducted experiments on pruning data with different proportions, where the pruned data were chosen randomly. This experiment helps us better understand how much data is required to power the model. We used the model trained with the dataset using background and marker substitution. We summarized the results in Tables \ref{prungTestS3} and \ref{prungTestS4}, and provided visual results in Fig. \ref{performanceAug1} and \ref{performanceAug2}. The tests on $S_3$ show that with 2\% of the data, the model achieves a precision of 0.882 in PE. With 4\% of the data, the precision exceeds 0.9, reaching 0.926. With 40\% of the data, the model achieves a precision of 0.993. The tests on $S_4$ show that with 2\% of the data, the model achieves a precision of 0.890 in PE, and with 6\% of the data, the precision exceeds 0.95, reaching 0.970. With 40\% of the data, the model achieves a precision of 0.991. These results indicate that approximately 4\% of our dataset can enable the model to achieve good precision under the designed data augmentation, which is approximately 120 images. However, the validation set is sampled separately and randomly, making it difficult to cover all edge cases. The data sampling procedure was targeted to achieve high precision and aimed to cover as many edge cases as possible to achieve robust results during real-time operation. Each batch sampling covers the error cases of the models trained with previous batches. High precision is crucial for the interactive system, especially for LS or LS training. Our semi-automated dataset generation method underwent 18 rounds during the data sampling stage, resulting in a robust dataset for reliable inference under our system setup.

\section{Conclusion}
In this work, we introduced a novel approach to extend the capabilities of the EasyVis system, bringing it closer to application in real surgery by enabling real-time multi-view 3D reconstruction for markerless surgical tools and operation in complex environments, including real tissue as the background. We proposed a novel LS tool pose dataset to fill the gap in existing surgical tool datasets. A training strategy was developed to enable multi-view consistent keypoint estimation without markers, achieving good precision in 3D reconstruction. A semi-automated dataset generation method was developed to create the ST-Pose dataset, which was used to train a YOLOv8-Pose model and deploy it within the EasyVis framework. Through iterative dataset generation, we achieved high accuracy during validation under the EasyVis setup and LS box trainer bean drop task setup, with a precision of 0.993 in surgical tool 2D pose estimation, a BPE\_PD score of 7.689 for 3D reconstruction quality, and an LPIPS of 0.106 for 3D rendering quality. This precision indicates that our work is robust and precise against variations within our LS box trainer setup. Visual experiment results of back-projecting reconstructed 3D space to existing observation views demonstrate the precision of our virtual 3D scene reconstruction for the LS box trainer. Further experiments on animal tissue suggest that our work has potential for real surgical applications. When maneuvering the surgical tool over animal tissue with observations from five top views, our system can render any novel view, including side views, providing a direct visual sense of depth and enabling the determination of distances between surgical tools and tissue. The average operation time is 12.6 ms per frame, meeting real-time performance requirements and allowing room for further algorithm development. In summary, this work is a step toward real-time 3D visualization for real laparoscopic surgery.



\bibliographystyle{ieeetr} 
\bibliography{template}

\begin{thebibliography}{10}

\bibitem{article}
V.~Penza, J.~Ortiz, L.~Mattos, A.~Forgione, and E.~De~Momi, ``Dense soft tissue 3d reconstruction refined with super-pixel segmentation for robotic abdominal surgery,'' {\em International journal of computer assisted radiology and surgery}, vol.~11, 09 2015.

\bibitem{inproceedings}
N.~Mahmoud, I.~Cirauqui, A.~Hostettler, C.~Doignon, L.~Soler, J.~Marescaux, and J.~Montiel, ``Orbslam-based endoscope tracking and 3d reconstruction,'' vol.~10170, pp.~72--83, 02 2017.

\bibitem{10.1007/978-3-540-75757-3_9}
M.~Hu, G.~Penney, P.~Edwards, M.~Figl, and D.~J. Hawkes, ``3d reconstruction of internal organ surfaces for minimal invasive surgery,'' in {\em Medical Image Computing and Computer-Assisted Intervention -- MICCAI 2007} (N.~Ayache, S.~Ourselin, and A.~Maeder, eds.), (Berlin, Heidelberg), pp.~68--77, Springer Berlin Heidelberg, 2007.

\bibitem{10.1007/978-3-031-16449-1_41}
Y.~Wang, Y.~Long, S.~H. Fan, and Q.~Dou, ``Neural rendering for stereo 3d reconstruction of deformable tissues in robotic surgery,'' in {\em Medical Image Computing and Computer Assisted Intervention -- MICCAI 2022} (L.~Wang, Q.~Dou, P.~T. Fletcher, S.~Speidel, and S.~Li, eds.), (Cham), pp.~431--441, Springer Nature Switzerland, 2022.

\bibitem{son2023advancements}
G.~M. Son, ``Advancements and challenges in minimally invasive surgery training among general-surgery residents in thailand,'' {\em Journal of Minimally Invasive Surgery}, vol.~26, no.~4, p.~178, 2023.

\bibitem{grossberg2021conscious}
S.~Grossberg, {\em Conscious mind, resonant brain: how each brain makes a mind}.
\newblock Oxford University Press, 2021.

\bibitem{katz2022dual}
J.~Katz, H.~Hua, S.~Lee, M.~Nguyen, and A.~Hamilton, ``A dual-view multi-resolution laparoscope for safer and more efficient minimally invasive surgery,'' {\em Scientific Reports}, vol.~12, no.~1, p.~18444, 2022.

\bibitem{ke2020towards}
J.~Ke, A.~J. Watras, J.-J. Kim, H.~Liu, H.~Jiang, and Y.~H. Hu, ``Towards real-time, multi-view video stereopsis,'' in {\em ICASSP 2020-2020 IEEE International Conference on Acoustics, Speech and Signal Processing (ICASSP)}, pp.~1638--1642, IEEE, 2020.

\bibitem{ackerman2002surface}
J.~D. Ackerman, K.~Keller, and H.~Fuchs, ``Surface reconstruction of abdominal organs using laparoscopic structured light for augmented reality,'' in {\em Three-Dimensional Image Capture and Applications V}, vol.~4661, pp.~39--46, SPIE, 2002.

\bibitem{maurice2012structured}
X.~Maurice, C.~Albitar, C.~Doignon, and M.~de~Mathelin, ``A structured light-based laparoscope with real-time organs' surface reconstruction for minimally invasive surgery,'' in {\em 2012 Annual International Conference of the IEEE Engineering in Medicine and Biology Society}, pp.~5769--5772, IEEE, 2012.

\bibitem{clancy2015dual}
N.~T. Clancy, J.~Lin, S.~Arya, G.~B. Hanna, and D.~S. Elson, ``Dual multispectral and 3d structured light laparoscope,'' in {\em Multimodal Biomedical Imaging X}, vol.~9316, pp.~60--64, SPIE, 2015.

\bibitem{reiter2014surgical}
A.~Reiter, A.~Sigaras, D.~Fowler, and P.~K. Allen, ``Surgical structured light for 3d minimally invasive surgical imaging,'' in {\em 2014 IEEE/RSJ International Conference on Intelligent Robots and Systems}, pp.~1282--1287, IEEE, 2014.

\bibitem{ghahremani2020ffd}
M.~Ghahremani, Y.~Liu, and B.~Tiddeman, ``Ffd: Fast feature detector,'' {\em IEEE Transactions on Image Processing}, vol.~30, pp.~1153--1168, 2020.

\bibitem{bouguet1995proceedings}
J.~Bouguet and P.~Perona, ``Proceedings of the international conference on computer vision,'' 1995.

\bibitem{rosten2005fusing}
E.~Rosten and T.~Drummond, ``Fusing points and lines for high performance tracking,'' in {\em Tenth IEEE International Conference on Computer Vision (ICCV'05) Volume 1}, vol.~2, pp.~1508--1515, Ieee, 2005.

\bibitem{mildenhall2021nerf}
B.~Mildenhall, P.~P. Srinivasan, M.~Tancik, J.~T. Barron, R.~Ramamoorthi, and R.~Ng, ``Nerf: Representing scenes as neural radiance fields for view synthesis,'' {\em Communications of the ACM}, vol.~65, no.~1, pp.~99--106, 2021.

\bibitem{sun2025easyvis}
Y.-H. Sun, J.~Ke, J.~Fernandes, J.~Chen, H.~Jiang, and Y.~H. Hu, ``Easyvis: a real-time 3d visualization software system for laparoscopic surgery box trainer,'' {\em Updates in Surgery}, pp.~1--16, 2025.

\bibitem{yolov8}
R.~Varghese and S.~M., ``Yolov8: A novel object detection algorithm with enhanced performance and robustness,'' in {\em 2024 International Conference on Advances in Data Engineering and Intelligent Computing Systems (ADICS)}, pp.~1--6, 2024.

\bibitem{algiriyage2021towards}
N.~Algiriyage, R.~Prasanna, K.~Stock, E.~Hudson-Doyle, D.~Johnston, M.~Punchihewa, and S.~Jayawardhana, ``Towards real-time traffic flow estimation using yolo and sort from surveillance video footage.,'' in {\em ISCRAM}, pp.~40--48, 2021.

\bibitem{qureshi2023comprehensive}
R.~Qureshi, M.~G. RAGAB, S.~J. ABDULKADER, A.~ALQUSHAIB, E.~H. SUMIEA, H.~Alhussian, {\em et~al.}, ``A comprehensive systematic review of yolo for medical object detection (2018 to 2023),'' {\em Authorea Preprints}, 2023.

\bibitem{xu2024surgical}
Z.~Xu, M.~Yu, F.~Chen, H.~Wu, and F.~Luo, ``Surgical tool detection in open surgery based on faster r-cnn, yolo v5 and yolov8,'' in {\em 2024 IEEE 7th Advanced Information Technology, Electronic and Automation Control Conference (IAEAC)}, pp.~1830--1834, IEEE, 2024.

\bibitem{almufareh2024automated}
M.~F. Almufareh, M.~Imran, A.~Khan, M.~Humayun, and M.~Asim, ``Automated brain tumor segmentation and classification in mri using yolo-based deep learning,'' {\em IEEE Access}, 2024.

\bibitem{maji2022yolo}
D.~Maji, S.~Nagori, M.~Mathew, and D.~Poddar, ``Yolo-pose: Enhancing yolo for multi person pose estimation using object keypoint similarity loss,'' in {\em Proceedings of the IEEE/CVF Conference on Computer Vision and Pattern Recognition}, pp.~2637--2646, 2022.

\bibitem{rahati2022sports}
A.~Rahati and K.~Rahbar, ``Sports movements modification based on 2d joint position using yolo to 3d skeletal model adaptation,'' {\em Journal of AI and Data Mining}, vol.~10, no.~4, pp.~549--557, 2022.

\bibitem{sznitman2012data}
R.~Sznitman, K.~Ali, R.~Richa, R.~H. Taylor, G.~D. Hager, and P.~Fua, ``Data-driven visual tracking in retinal microsurgery,'' in {\em Medical Image Computing and Computer-Assisted Intervention--MICCAI 2012: 15th International Conference, Nice, France, October 1-5, 2012, Proceedings, Part II 15}, pp.~568--575, Springer, 2012.

\bibitem{maier2014can}
L.~Maier-Hein, S.~Mersmann, D.~Kondermann, S.~Bodenstedt, A.~Sanchez, C.~Stock, H.~G. Kenngott, M.~Eisenmann, and S.~Speidel, ``Can masses of non-experts train highly accurate image classifiers? a crowdsourcing approach to instrument segmentation in laparoscopic images,'' in {\em Medical Image Computing and Computer-Assisted Intervention--MICCAI 2014: 17th International Conference, Boston, MA, USA, September 14-18, 2014, Proceedings, Part II 17}, pp.~438--445, Springer, 2014.

\bibitem{zhao20206d}
W.~Zhao, S.~Zhang, Z.~Guan, H.~Luo, L.~Tang, J.~Peng, and J.~Fan, ``6d object pose estimation via viewpoint relation reasoning,'' {\em Neurocomputing}, vol.~389, pp.~9--17, 2020.

\bibitem{zhu2021aspp}
Y.~Zhu, L.~Wan, W.~Xu, and S.~Wang, ``Aspp-df-pvnet: atrous spatial pyramid pooling and distance-filtered pvnet for occlusion resistant 6d object pose estimation,'' {\em Signal Processing: Image Communication}, vol.~95, p.~116268, 2021.

\bibitem{wang2021gdr}
G.~Wang, F.~Manhardt, F.~Tombari, and X.~Ji, ``Gdr-net: Geometry-guided direct regression network for monocular 6d object pose estimation,'' in {\em Proceedings of the IEEE/CVF Conference on Computer Vision and Pattern Recognition}, pp.~16611--16621, 2021.

\bibitem{wen2024foundationpose}
B.~Wen, W.~Yang, J.~Kautz, and S.~Birchfield, ``Foundationpose: Unified 6d pose estimation and tracking of novel objects,'' in {\em Proceedings of the IEEE/CVF Conference on Computer Vision and Pattern Recognition}, pp.~17868--17879, 2024.

\bibitem{9197555}
M.~Tian, L.~Pan, M.~H. Ang, and G.~Hee~Lee, ``Robust 6d object pose estimation by learning rgb-d features,'' in {\em 2020 IEEE International Conference on Robotics and Automation (ICRA)}, pp.~6218--6224, 2020.

\bibitem{song2024rgb}
Y.~Song and C.~Tang, ``A rgb-d feature fusion network for occluded object 6d pose estimation,'' {\em Signal, Image and Video Processing}, pp.~1--11, 2024.

\bibitem{hartley2003multiple}
R.~Hartley and A.~Zisserman, {\em Multiple view geometry in computer vision}.
\newblock Cambridge university press, 2003.

\bibitem{toshev2014deeppose}
A.~Toshev and C.~Szegedy, ``Deeppose: Human pose estimation via deep neural networks,'' in {\em Proceedings of the IEEE conference on computer vision and pattern recognition}, pp.~1653--1660, 2014.

\bibitem{openpose}
Z.~{Cao}, G.~{Hidalgo Martinez}, T.~{Simon}, S.~{Wei}, and Y.~A. {Sheikh}, ``Openpose: Realtime multi-person 2d pose estimation using part affinity fields,'' {\em IEEE Transactions on Pattern Analysis and Machine Intelligence}, 2019.

\bibitem{pishchulin2016deepcut}
L.~Pishchulin, E.~Insafutdinov, S.~Tang, B.~Andres, M.~Andriluka, P.~V. Gehler, and B.~Schiele, ``Deepcut: Joint subset partition and labeling for multi person pose estimation,'' in {\em Proceedings of the IEEE conference on computer vision and pattern recognition}, pp.~4929--4937, 2016.

\bibitem{sun2019deep}
K.~Sun, B.~Xiao, D.~Liu, and J.~Wang, ``Deep high-resolution representation learning for human pose estimation,'' in {\em Proceedings of the IEEE/CVF conference on computer vision and pattern recognition}, pp.~5693--5703, 2019.

\bibitem{alphapose}
H.-S. Fang, J.~Li, H.~Tang, C.~Xu, H.~Zhu, Y.~Xiu, Y.-L. Li, and C.~Lu, ``Alphapose: Whole-body regional multi-person pose estimation and tracking in real-time,'' {\em IEEE Transactions on Pattern Analysis and Machine Intelligence}, 2022.

\bibitem{fang2017rmpe}
H.-S. Fang, S.~Xie, Y.-W. Tai, and C.~Lu, ``{RMPE}: Regional multi-person pose estimation,'' in {\em ICCV}, 2017.

\bibitem{guler2018densepose}
R.~A. G{\"u}ler, N.~Neverova, and I.~Kokkinos, ``Densepose: Dense human pose estimation in the wild,'' in {\em Proceedings of the IEEE conference on computer vision and pattern recognition}, pp.~7297--7306, 2018.

\bibitem{lomonaco2017core50}
V.~Lomonaco and D.~Maltoni, ``Core50: a new dataset and benchmark for continuous object recognition,'' in {\em Conference on robot learning}, pp.~17--26, PMLR, 2017.

\bibitem{chao2021dexycb}
Y.-W. Chao, W.~Yang, Y.~Xiang, P.~Molchanov, A.~Handa, J.~Tremblay, Y.~S. Narang, K.~Van~Wyk, U.~Iqbal, S.~Birchfield, {\em et~al.}, ``Dexycb: A benchmark for capturing hand grasping of objects,'' in {\em Proceedings of the IEEE/CVF Conference on Computer Vision and Pattern Recognition}, pp.~9044--9053, 2021.

\bibitem{andriluka2018posetrack}
M.~Andriluka, U.~Iqbal, E.~Insafutdinov, L.~Pishchulin, A.~Milan, J.~Gall, and B.~Schiele, ``Posetrack: A benchmark for human pose estimation and tracking,'' in {\em Proceedings of the IEEE conference on computer vision and pattern recognition}, pp.~5167--5176, 2018.

\bibitem{hodan2017t}
T.~Hodan, P.~Haluza, {\v{S}}.~Obdr{\v{z}}{\'a}lek, J.~Matas, M.~Lourakis, and X.~Zabulis, ``T-less: An rgb-d dataset for 6d pose estimation of texture-less objects,'' in {\em 2017 IEEE Winter Conference on Applications of Computer Vision (WACV)}, pp.~880--888, IEEE, 2017.

\bibitem{tremblay2018falling}
J.~Tremblay, T.~To, and S.~Birchfield, ``Falling things: A synthetic dataset for 3d object detection and pose estimation,'' in {\em Proceedings of the IEEE Conference on Computer Vision and Pattern Recognition Workshops}, pp.~2038--2041, 2018.

\bibitem{fabbri2021motsynth}
M.~Fabbri, G.~Bras{\'o}, G.~Maugeri, O.~Cetintas, R.~Gasparini, A.~O{\v{s}}ep, S.~Calderara, L.~Leal-Taix{\'e}, and R.~Cucchiara, ``Motsynth: How can synthetic data help pedestrian detection and tracking?,'' in {\em Proceedings of the IEEE/CVF International Conference on Computer Vision}, pp.~10849--10859, 2021.

\bibitem{xiang2020sapien}
F.~Xiang, Y.~Qin, K.~Mo, Y.~Xia, H.~Zhu, F.~Liu, M.~Liu, H.~Jiang, Y.~Yuan, H.~Wang, {\em et~al.}, ``Sapien: A simulated part-based interactive environment,'' in {\em Proceedings of the IEEE/CVF conference on computer vision and pattern recognition}, pp.~11097--11107, 2020.

\bibitem{ullman1979interpretation}
S.~Ullman, ``The interpretation of structure from motion,'' {\em Proceedings of the Royal Society of London. Series B. Biological Sciences}, vol.~203, no.~1153, pp.~405--426, 1979.

\bibitem{NVIDIA_TensorRT_2021}
NVIDIA, ``Tensorrt,'' March 2021.
\newblock [Online].

\bibitem{Cutie}
H.~K. Cheng, S.~W. Oh, B.~Price, J.-Y. Lee, and A.~Schwing, ``Putting the object back into video object segmentation,'' {\em arXiv preprint arXiv:2310.12982}, 2023.

\bibitem{lin2014microsoft}
T.-Y. Lin, M.~Maire, S.~Belongie, J.~Hays, P.~Perona, D.~Ramanan, P.~Doll{\'a}r, and C.~L. Zitnick, ``Microsoft coco: Common objects in context,'' in {\em Computer Vision--ECCV 2014: 13th European Conference, Zurich, Switzerland, September 6-12, 2014, Proceedings, Part V 13}, pp.~740--755, Springer, 2014.

\bibitem{furukawa2015multi}
Y.~Furukawa and C.~Hern{\'a}ndez, ``Multi-view stereo: A tutorial,'' {\em Foundations and Trends{\textregistered} in Computer Graphics and Vision}, vol.~9, no.~1-2, pp.~1--148, 2015.

\end{thebibliography}

\end{document}